\pdfoutput=1

\documentclass[11pt]{article}

\usepackage[]{acl}

\usepackage{times}
\usepackage{latexsym}

\usepackage[T1]{fontenc}

\usepackage[utf8]{inputenc}

\usepackage{microtype}

\usepackage{graphicx}
\usepackage{multirow}
\usepackage{booktabs}
\usepackage{amssymb}
\usepackage{subfigure}
\usepackage{bbding}
\usepackage{pifont}
\usepackage{amsmath}

\usepackage{algorithm,algpseudocode}
\algnewcommand{\algorithmicforeach}{\textbf{for each}}
\algdef{SE}[FOR]{ForEach}{EndForEach}[1]
  {\algorithmicforeach\ #1\ \algorithmicdo}
  {\algorithmicend\ \algorithmicforeach}

\title{Contextual Fine-to-Coarse Distillation for Coarse-grained Response Selection in Open-Domain Conversations}

\author{Wei Chen\textsuperscript{\rm 1}\thanks{~~Worked during the internship at Microsoft Research Asia. Zhongyu Wei and Yeyun Gong are corresponding authors.}, Yeyun Gong\textsuperscript{\rm 2},  Can Xu\textsuperscript{\rm 2}, Huang Hu\textsuperscript{\rm 2}, Bolun Yao\textsuperscript{\rm 3}, Zhongyu Wei\textsuperscript{\rm 1,4}, \\ \bf Zhihao Fan\textsuperscript{\rm 1}, Xiaowu Hu\textsuperscript{\rm 2}, Bartuer Zhou\textsuperscript{\rm 2}, Biao Cheng\textsuperscript{\rm 2}, Daxin Jiang\textsuperscript{\rm 2} and Nan Duan\textsuperscript{\rm 2} \\ \textsuperscript{\rm 1}School of Data Science, Fudan University, China \textsuperscript{\rm 2}Microsoft Research Asia, China \\ \textsuperscript{\rm 3}Nanjing University of Science and Technology, China \\ \textsuperscript{\rm 4}Research Institute of Intelligent and Complex Systems, Fudan University, China \\ \{chenwei18,zywei,fanzh18\}@fudan.edu.cn;\{yegong,caxu,huahu\}@microsoft.com,\\\{xiaowuhu, bazhou,bicheng,djiang,nanduan\}@microsoft.com;  yaobl001@njust.edu.cn}

\begin{document}
\maketitle
\begin{abstract}
We study the problem of coarse-grained response selection in retrieval-based dialogue systems. The problem is equally important with fine-grained response selection, but is less explored in existing literature. In this paper, we propose a \textbf{C}ontextual \textbf{F}ine-to-\textbf{C}oarse (CFC) distilled model for coarse-grained response selection in open-domain conversations. In our CFC model, dense representations of query, candidate contexts and responses is learned based on the multi-tower architecture using contextual matching, and richer knowledge learned from the one-tower architecture (fine-grained) is distilled into the multi-tower architecture (coarse-grained) to enhance the performance of the retriever. To evaluate the performance of the proposed model, we construct two new datasets based on the Reddit comments dump and Twitter corpus. Extensive experimental results on the two datasets show that the proposed method achieves huge improvement over all evaluation metrics compared with traditional baseline methods. 
\end{abstract}

\section{Introduction}
\label{section:intro}

\begin{figure}
\centering
\includegraphics[width=1.0\columnwidth]{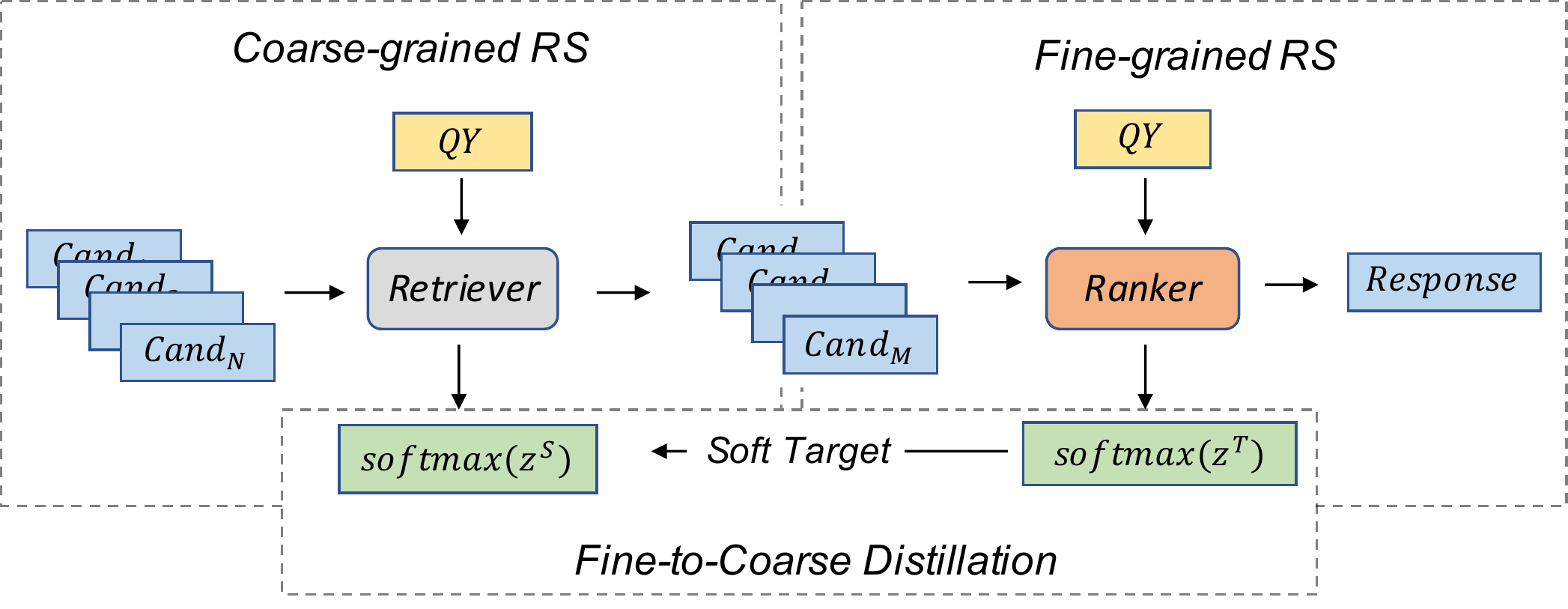}
\caption{A common structure of retrieval-based dialogue system, where coarse-grained RS provides a much smaller ($M\ll N$) candidate set for fine-grained RS.
\emph{QY} and \emph{Cand} are the abbreviations of \emph{query} and \emph{candidate} respectively.}
\label{fig:intro}
\end{figure}

Given utterances of a query, the retrieval-based dialogue (RBD) system aims to search for the most relevant response from a set of historical records of conversations \cite{higashinaka2014towards, yan2016learning, boussaha2019deep}. A complete RBD system usually contain two stages: coarse-grained response selection (RS) and fine-grained response selection \cite{fu2020context}. As shown in Figure \ref{fig:intro}, in coarse-grained RS stage, the retriever identifies a much smaller list of candidates (usually dozens) from large-scale candidate database (up to millions or more), then the ranker in fine-grained RS stage selects the best response from the retrieved candidate list.

Recent studies \cite{whang2020response, xu2020learning, xu2021topic, whang2021response} pay more attention on fine-grained RS and various complex models are proposed to compute the similarities between the query and candidates for response selection. Although promising improvements have been reported, the performance of fine-grained stage is inevitably limited by the quality of the candidate list constructed. Therefore, a high-quality coarse-grained RS module is crucial, which is less explored in existing literature \cite{lan2020ultra}.

In this paper, we focus on the task of coarse-grained response selection, i.e., dialogue response retrieval. There are two major challenges. First, different from general text matching tasks such as ad-hoc retrieval \cite{hui2018co} or question answering (QA) retrieval \cite{karpukhin2020dense}, keywords overlapping between context and response in dialogue are potentially rare, such as when a topic transition \cite{sevegnani2021otters} occurs in response. This makes it difficult to directly match the query with candidate responses. Second, compared with fine-grained RS, coarse-grained RS deals with much larger number of candidates. Therefore, it is impractical to apply complex matching model that jointly process query and response for the similarity computation like in fine-grained RS, due to the retrieval latency (traverse millions of candidates online). Instead, the efficient BM25 system \cite{robertson2009probabilistic} based on sparse representations is the mainstream algorithm in coarse-grained text matching. 

To mitigate the above mentioned two problems, we propose a \textbf{C}ontextual \textbf{F}ine-to-\textbf{C}oarse (CFC) distilled model for coarse-grained RS. Instead of matching query with response directly, we propose a novel task of query-to-context matching in coarse-grained retrieval, i.e. \emph{contextual matching}. Given a query, it is matched with candidate contexts to find most similar ones, and the corresponding responses are returned as the retrieved result. In this case, the potential richer keywords in the contexts can be utilized. To take the advantage of complex model and keep the computation cost acceptable, we distillate the knowledge learned from fine-grained RS into coarse-grained RS while maintaining the original architecture. 

For the evaluation, there is no existing dataset that can be used to evaluate our model in the setting of contextual matching, because it needs to match context with context during training, while positive pairs of context-context is not naturally available like context-response pairs. Therefore, we construct two datasets based on Reddit comment dump and Twitter corpus. Extensive experimental results show that our proposed model greatly improve the retrieval recall rate and the perplexity and relevance of the retrieved responses on both datasets.


The main contributions of this paper are three-fold: 1) We explore the problem of coarse-grained RS in open domain conversations and propose a Contextual Fine-to-Coarse (CFC) distilled model; 2) We construct two new datasets based on Reddit comment dump and Twitter corpus, as a new benchmark to evaluate coarse-grained RS task; 3) We construct extensive experiments to demonstrate the effectiveness and potential of our proposed model in coarse-grained RS. 


\section{Related Work}
\label{section:related-work}

\paragraph{Fine-grained Response Selection} ~ {In recent years, many works have been proposed to improve the performance of fine-grained selection module in retrieval-based chatbots \cite{zhang2018modeling,zhou2018multi,tao2019multi,whang2019domain,yuan2019multi}. Owing to the rapid development of pre-trained language models (PLMs) \cite{radford2019language}, recent works \cite{gu2020speaker, whang2021response, sevegnani2021otters} achieve the state-of-the-art (SOTA) results by utilizing PLMs such as BERT \cite{devlin2018bert} to model cross-attention and complex intersection between the context and response.}

\paragraph{Coarse-grained Response Selection} ~ {On the other hand, coarse-grained dialogue retrieval is an important but rarely explored field. Limited by efficiency, there are usually two methods for coarse-grained response selection, i.e., the sparse representations based method represented by BM25 \cite{robertson2009probabilistic}, and the dense representations based method represented by dual-Encoder \cite{chidambaram2018learning, humeau2019poly, karpukhin2020dense, lan2020ultra, lin2020distilling}.}

\section{Method}
\label{section:method}

\begin{figure*}
\centering
\subfigure[Two-tower model based on QS matching]{
\begin{minipage}[b]{0.35\textwidth}
\includegraphics[width=1\textwidth]{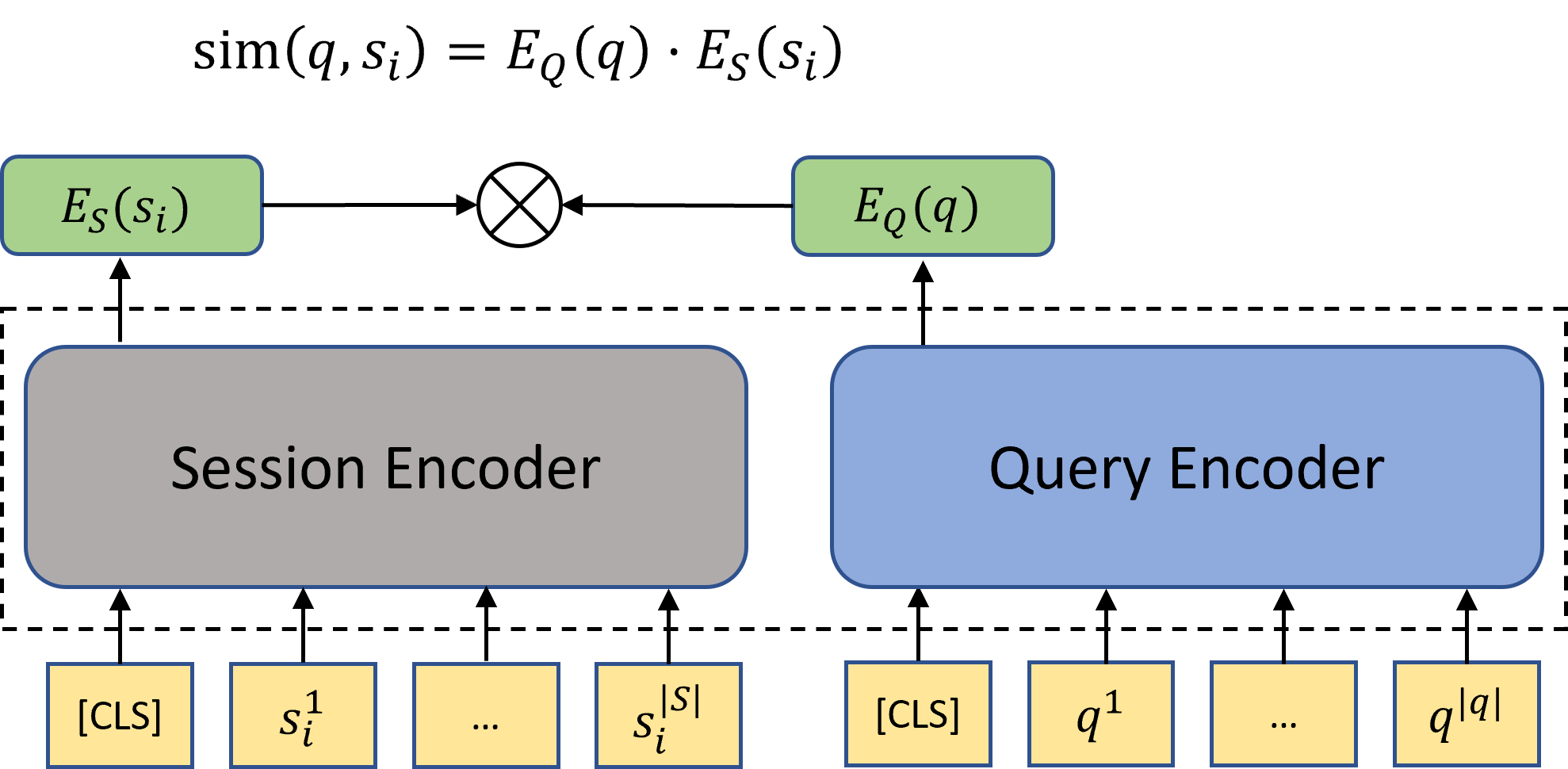}
\end{minipage}
\label{fig:two_tower}}
\subfigure[Three-tower model based on DQS matching]{
\begin{minipage}[b]{0.55\textwidth}
\includegraphics[width=1\textwidth]{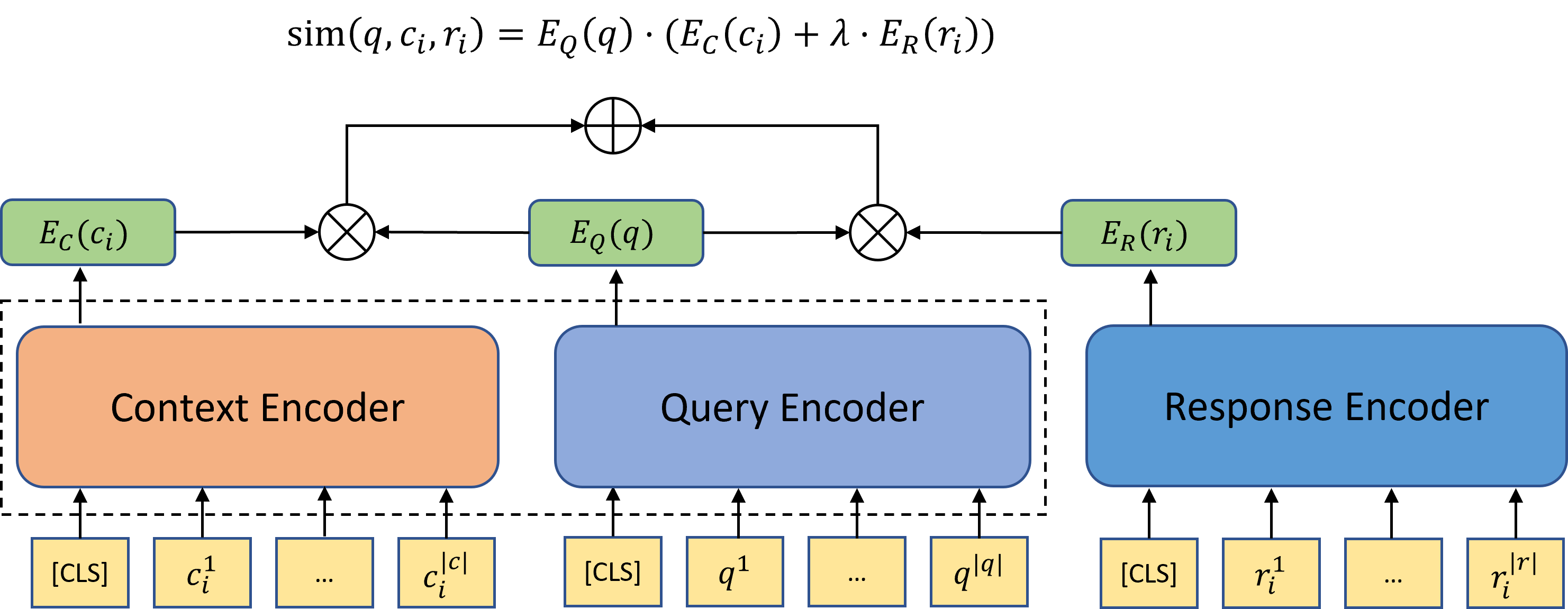}
\end{minipage}
\label{fig:three_tower}}
\caption{Multi-tower architecture with independent encoders, the hidden representation of the [CLS] token of each sequence is passed through a linear layer followed by a hyperbolic tangent (Tanh) activation function to get the dense representations (embeddings) of the entire sentence.}
\label{fig:model}
\end{figure*}



In coarse-grained response selection, there is a fixed candidate database containing a large number of \emph{context-response} pairs. Formally, given a \emph{query}, i.e., a new context, the goal is to retrieve Top-K most suitable \emph{responses} for the \emph{query} from the candidate database.

We propose a contextual fine-to-coarse distillation framework for the task of coarse-grained RS. First, we formulate the problem as a task of \textbf{contextual matching}, i.e., match query with context instead response; Second, we utilize a \textbf{multi-tower architecture} to deal with the similarity computation of query and candidates in contextual matching; Third, we utilize \textbf{knowledge distillation} to leverage the deep interaction between query and response learned in one-tower architecture. 

\subsection{Contextual Matching}

An intuitive idea of coarse-grained RS is to treat all responses as candidate documents and directly use query to retrieve them, while this non-contextual approach results in a quite low retrieval recall rate \cite{lan2020ultra}. Inspired by recent studies of context-to-context matching in fine-grained RS \cite{fu2020context}, we propose contextual matching in coarse-grained RS, which is to match the query with candidate contexts, and return the responses corresponding to the most similar contexts. We consider three ways of contextual matching.

\paragraph{Query-Context (QC)} ~ {In QC matching, we treat contexts instead of responses as candidate documents. At run-time, we calculate the similarities between query and candidate contexts, and the responses corresponding to the Top-K most similar contexts are returned as the retrieved results. The motivation of using QC matching is similar contexts may also share similar responses.}

\paragraph{Query-Session (QS)} ~ {A \textbf{session} represents the concatenated text of context and corresponding response \cite{fu2020context}, which we think is more informative than context alone. In QS matching, we treat sessions as candidate documents and return the responses in Top-K most similar sessions as the retrieved results.}

\paragraph{Decoupled Query-Session (DQS)} ~ {Apart from QS matching, we also consider a decoupled way to match query with candidate sessions. In DQS matching, we treat contexts and responses as independent candidate documents. Similarities between query and contexts, query and responses are first calculated independently, then the query-session similarity can be obtained by the weighted sum. QS and DQS matching are actually two different ways to calculate query-session similarity.}

\subsection{Multi-Tower Architecture}




For the retriever to search large-scale candidates with low latency, neural-based retrievers are usually designed as (or limited to) \emph{multi-tower} architecture (Figure \ref{fig:model}). In multi-tower models, the query and the candidates are independently mapped to a common vector space by different encoders, where similarity can be calculated. After training, the embeddings of large-scale candidates can be \textbf{pre-calculated offline}, and only the embedding of query needs to be calculated online. In this way, fast sublinear-time approximation methods such as approximate nearest neighbor search \cite{shrivastava2014asymmetric} can be utilized to search for Top-K vectors that are most similar to the query, which can achieve an acceptable retrieval latency during inference. 






\subsubsection{Two-Tower Model}

For QC and QS matching, \emph{two-tower} architecture is adopted. Taking QS matching as an example (Figure \ref{fig:two_tower}), the dense \emph{session encoder} $E_{S}(\cdot)$ maps any candidate session to real-valued embedding vectors in a $d$-dimensional space, and an index is built for all the $N$ session vectors for retrieval. At run-time, a different dense \emph{query encoder} $E_{Q}(\cdot)$ maps the query to a $d$-dimensional vector, and retrieves $k$ candidate sessions of which vectors are the closest to the query vector. We use the \emph{dot product} of vectors as the similarity between query and candidate session following \cite{karpukhin2020dense}. 

\subsubsection{Three-Tower Model}

For DQS matching, dense representations of query, context and response are independently calculated, the architecture is thus designed as \emph{three-tower} with three encoders, which is \emph{query encoder} $E_{Q}(\cdot)$, \emph{context encoder} $E_{C}(\cdot)$ and \emph{response encoder} $E_{R}(\cdot)$ (Figure \ref{fig:three_tower}). Similarly, context and response vectors are calculated and cached offline respectively and two indexes are built for retrieving them. The final similarity of query and session is weighted by the dot product of query-context and query-response. The weighting coefficient $\lambda$ can be adjusted to determine whether it is biased to match the context or match the response\footnote{In all experiments in this paper, $\lambda$ is set to 1 to treat candidate context and response equally.}. 

\subsubsection{Training Multi-Tower Model}

We unify the training of the two-tower and three-tower models by formalizing them into a same \emph{metric learning} problem \cite{kulis2012metric}. The goal is to learn a matching space where similarities between positive pairs is higher than negative ones, by learning a better embedding function. We use the training of three-tower model (DQS matching) as an example. Formally, we denote the training set as $\mathcal{D} = \{q_{i}, \{k_{i}^{+}, k_{i}^{-}\}\}_{i=1}^N$. Each training instance contains a query $q_{i}$, a set of positive examples $k_{i}^{+}$ and a set of negative examples $k_{i}^{-}$. Among them, $k_{i}^{+}$ contain several positive contexts and several positive responses, similarly, $k_{i}^{-}$ contain several negative contexts and several negative responses. We optimize the loss function as the sum of negative log likelihood of all positive pairs simultaneously:

\begin{equation}
\begin{aligned}
\mathcal{L}(q_{i}) = -log\frac{\sum_{k^{'} \in \{k_{i}^{+}\}}e^{{\rm sim}(q_{i}, k^{'})}}{\sum_{k^{'} \in \{k_{i}^{+}, k_{i}^{-}\}}e^{{\rm sim}(q_{i}, k^{'})}}
\end{aligned}
\label{equation:contrastive_learning}
\end{equation}

where the similarity function is defined as: 

\begin{equation}
{\rm sim}(q_{i}, k^{'}) = E_{Q}(q_{i}) \cdot E(k^{'}).
\label{equation:sim}
\end{equation}

The embedding function $E(\cdot)$ of $k^{'}$ in Equation \ref{equation:sim} can be $E_{C}(\cdot)$ or $E_{R}(\cdot)$, depending on the type of $k^{'}$. 

\paragraph{Positive and negative examples} ~ {The core issue of training multi-tower models for contextual matching is to find positive pairs of query-context (or query-session). In this paper, we assume that contexts with exactly the same response are positive samples of each other, which is a cautious but reliable strategy. Formally, given a response $r$, if there are multiple contexts whose response is $r$, then we can randomly selected one context as the query $q$, and the other contexts are \textbf{positive contexts} of $q$, and $r$ is the \textbf{positive response} of $q$. Negative samples of contexts and responses can be obtained from in-batch \cite{karpukhin2020dense} or random sampling from database. Similarly, positive query-session is obtained by replacing the context in positive query-context with the whole session.}

\subsection{Distillation from One-Tower Model}

In multi-tower architecture, the query and candidates are expressed by their embeddings independently, which may cause the loss of information, and their monotonous way of interaction (inner product) further limits the capability \cite{lin2020distilling}. Comparing with multi-tower model, one-tower model takes both the query and the candidate as a concatenated input and allow the cross attention between query and candidate in self-attention layer. Despite fewer parameters, one-tower model have been shown to learn a more informative representations than multi-tower model, thus it is preferred in fine-grained RS \cite{yang2020retriever}. To leverage the richer expressiveness learned by the one-tower model, knowledge from one-tower model is distilled into multi-tower model to enhance the retriever.

\subsubsection{Training One-Tower Model}


Before distillation, we need to train teacher models based on one-tower architecture. Let's take the training of teacher model for QS matching as an example. A single encoder is trained to distinguish whether the query and the session are relevant (positive), and the form is exactly same as the next sentence prediction (NSP) task in the BERT \cite{devlin2018bert} pre-training. Formally, given a training set $\mathcal{D} = \{q_{i}, s_{i}, l_{i}\}_{i=1}^N$, where $q_{i}$ is the query, $s_{i}$ is the candidate session and $l_{i} \in \{0, 1\}$ denotes whether $q_{i}$ and $s_{i}$ is a positive pair. To be specific, given a query $q$ and candidate session $s$, the encoder obtains the joint representation of the concatenated text of $q$ and $s$, and then computes the similarity score through a linear layer, the training objective is binary cross entropy loss. 


We summarize the \textbf{main difference} between one-tower and multi-tower as follows: one-tower model is more expressive, but less efficient and cannot handle large-scale candidates. The main reason is that feature-based method of calculating similarity scores rather than inner product limits the capability of offline caching. For new queries, the similarities with all candidates can only be calculated by traversal. The huge latency makes it impossible to use one-tower model in coarse-grained response retrieval. To leverage the expressiveness of one-tower model, we propose fine-to-coarse distillation, which can learn the knowledge of one-tower model while keeping the multi-tower structure unchanged, thereby improving the performance of the retriever.

\subsubsection{Fine-to-Coarse Distillation}

Take the two-tower \textbf{student} model (denoted as $S$) for QS matching  as an example, suppose we have trained the corresponding one-tower \textbf{teacher} model (denoted as $T$). For a given query $q$, suppose there are a list of sessions $\{s^{+}, s_{1}^{-}, ..., s_{n}^{-}\}$ and the corresponding label $y = \{1, 0, ..., 0\} \in \mathcal{R}^{n+1}$, that is, one positive session and $n$ negative sessions. We denote the similarity score vector of query-sessions computed by student model $S$ (Equation \ref{equation:sim}) as $z^{S} \in \mathcal{R}^{n+1}$, then the objective of Equation \ref{equation:contrastive_learning} is equivalent to maximizing the Kullback–Leibler (KL) divergence \cite{van2014renyi} of the two distributions: ${\rm softmax}(z^S)$ and $y$, where ${\rm softmax}$ function turns the score vector to probability distribution. 



The one-hot label $y$ treats each negative sample equally, while the similarity between query with each negative sample is actually different. To learn more accurate labels, we further use teacher model $T$ to calculate the similarity score vector between $q$ and $S$, denoted as $z^{T} \in \mathcal{R}^{n+1}$. We then replace the original training objective with minimizing KL divergence of the two distributions ${\rm softmax}(z^{S})$ and ${\rm softmax}(z^{T})$ (Figure \ref{fig:intro}), where the temperature parameter is applied in ${\rm softmax}$ function to avoid saturation.

The method of fine-to-coarse distillation is to push the student model (multi-tower) to learn the predicted label of teacher model (one-tower) as a soft target instead of original one-hot label. By fitting the label predicted by the teacher model, the multi-tower model can learn a more accurate similarity score distribution from the one-tower model while keeping the structure unchanged.




\section{Datasets Construction}
\label{section:data}


To evaluate the performance of the proposed model, we construct two new datasets based on the Reddit comments dump \cite{zhang2019dialogpt} and Twitter corpus\footnote{\url{https://github.com/Marsan-Ma-zz/chat_corpus}}. We create a training set, a multi-contexts (MC) test set and a candidate database for Reddit and Twitter respectively. For Reddit, we create an additional single-context (SC) test set. The motivation for these settings is explained in \S~\ref{twitter}. The size of our candidate database is one million in Twitter and ten million in Reddit respectively, which is very challenging for response retrieval. Table \ref{tab:data-static} shows the detailed statistics. We use exactly the same steps to build dataset for Reddit and Twitter, and similar datasets can also build from other large dialogue corpus in this way. 


\paragraph{MC test set} ~ {We first find out a set of responses with multiple contexts from candidate database, denoted as $R$. For each response $r$ in $R$, we randomly select one context $c$ from its all corresponding contexts $C_{r}$ to construct a context-response (CR) pair, and put the others contexts (denoted as $C_{r}^{-}$) back to the database. Our MC test set consists of these CR pairs. Each response in MC test set has multiple contexts, which ensures that there exits other contexts in the database that also correspond to this response, so the retrieval recall rate can be computed to evaluate the MC test set.}

\paragraph{SC test set} ~ {We create another test set (SC) for Reddit dataset. Contrary to the MC test set, each response in SC test set has only one context, i.e., there is no context in the database that exactly corresponds to the response. Obviously, the retrieval recall rate is invalid (always zero) on SC test set. We introduce other methods to evaluate SC test set in \S~\ref{section:metric}. The SC test set is a supplement to the MC test set which can evaluate the quality of retrieved responses given those ``unique" contexts.}


\paragraph{Candidate database} ~ {To adapt to different retrieval methods, the candidate database is designed with 4 fields, namely \emph{context}, \emph{response}, \emph{session}. Our candidate database consists of random context-response pairs except those in the MC and SC test sets. Besides, as mentioned above, those unselected context-response pairs ($C_{r}^{-}$) are deliberately merged into the database.}






\paragraph{Train set} ~ {The construction of training set is intuitive and similar to test set. It consists of responses and their corresponding multiple contexts. Formally, the training set can be denote as $D = \{r_{i}, c_{i, 1}, ..., c_{i, q}\}_{i=1}^{N}$, $r_{i}$ is a response and $\{c_{i, 1}, ..., c_{i, q}\}$ are all contexts with response $r_{i}$, where $q$ depends on $r_{i}$, and $q \geq 2$.} 


It is worth noting that there is no overlap between the contexts in the database and the contexts in the training set, which may prevent potential data leakage during training process to overestimate the evaluation metrics. The details of dataset construction are introduced in Appendix \ref{appendix:data_detail}.


\begin{table}[]
\centering
\scalebox{0.8}{
\begin{tabular}{ccccc}
\toprule
\multirow{2}{*}{\textbf{Datasets}} & \multirow{2}{*}{\textbf{Training set}} & \multicolumn{2}{c}{\textbf{Test set}} & \multirow{2}{*}{\textbf{Database}} \\ 
        &      & \textbf{MC} & \textbf{SC} &        \\ \midrule
Reddit  & 300K & 20K         & 20K         & 10M \\
Twitter & 20K  & 2K          & -           & 1M    \\ \bottomrule
\end{tabular}}
\caption{Data statistics of our new constructed datasets.}
\label{tab:data-static}
\end{table}

\section{Experiments}
\label{section:experiment}

We conduct extensive experiments on the constructed datasets. In this section, we present experimental settings, evaluation metrics, model performance, human evaluation, etc. to demonstrate the effectiveness of the proposed models.

\begin{table*}[]
\centering
\scalebox{0.8}{
\begin{tabular}{ccccccccccccc}
                                                & \multicolumn{8}{c}{MC Test Set}                                                                                                                       & \multicolumn{4}{c}{SC Test Set}                                    \\ \toprule
\multicolumn{1}{l|}{\multirow{2}{*}{Retriever}} & \multicolumn{4}{c}{Coverage@K}                              & \multicolumn{2}{c}{Perplexity@K} & \multicolumn{2}{c|}{Relevance@K}                     & \multicolumn{2}{c}{Perplexity@K} & \multicolumn{2}{c}{Relevance@K} \\
\multicolumn{1}{c|}{}                           & Top-1        & Top-20       & Top-100       & Top-500       & Top-1           & Top-20         & Top-1          & \multicolumn{1}{c|}{Top-20}         & Top-1           & Top-20         & Top-1          & Top-20         \\ \hline
\multicolumn{1}{l|}{Gold}                       & -            & -            & -             & -             & \multicolumn{2}{c}{205.7}       & \multicolumn{2}{c|}{73.1}                            & \multicolumn{2}{c}{181.8}       & \multicolumn{2}{c}{82.0}        \\ \hline
\multicolumn{13}{l}{\textbf{Contextual matching}}  \\ \hline
\multicolumn{1}{l|}{BM25-QC}                    & 1.1          & 3.9          & 5.7           & 7.8           & 210.5           & 217.9          & 61.5          & \multicolumn{1}{c|}{53.5}          & 208.3           & 217.5          & 60.6          & 52.1          \\
\multicolumn{1}{l|}{BM25-QS}                    & 0.9          & 3.6          & 5.8           & 8.3           & 207.7           & 214.2          & 80.0          & \multicolumn{1}{c|}{73.9}          & 200.0           & 208.3          & 81.6          & 74.1          \\ \hline
\multicolumn{1}{l|}{BE-QC}                      & 1.3          & 5.3          & 8.1           & 12.3          & 205.4           & 211.5          & 81.3          & \multicolumn{1}{c|}{75.8}          & 194.4           & 203.2          & 82.9          & 78.3          \\
\multicolumn{1}{l|}{BE-QS}                      & 1.6          & 5.9          & 11.8          & 20.4          & 200.1           & 206.1          & 85.0          & \multicolumn{1}{c|}{80.2}          & 190.9           & 199.8          & 85.3          & 80.6          \\ 
\multicolumn{1}{l|}{TE-DQS}                     & 1.5          & 5.5          & 9.7           & 18.1          & 201.3           & 207.5          & 84.8          & \multicolumn{1}{c|}{79.8}          & 190.5           & 198.2          & 85.5          & 80.4          \\ \hline
\multicolumn{1}{l|}{CFC-QC}                     & 2.9          & 6.5          & 9.1           & 13.0          & 199.5           & 208.9          & 84.9          & \multicolumn{1}{c|}{78.6}          & 187.5           & 196.3          & 86.2          & 80.8          \\
\multicolumn{1}{l|}{CFC-QS}                     & \textbf{4.2} & \textbf{7.8} & \textbf{13.1} & \textbf{21.3} & \textbf{194.8}  & \textbf{203.1} & \textbf{87.8} & \multicolumn{1}{c|}{\textbf{82.8}} & \textbf{184.3}  & 193.1          & \textbf{88.3} & \textbf{83.4} \\
\multicolumn{1}{l|}{CFC-DQS}                 & 3.7          & 7.3          & 12.7          & 19.4          & 196.5           & 205.3          & 86.9          & \multicolumn{1}{c|}{81.9}          & 184.8           & \textbf{192.6} & 88.1          & 83.3          \\ \hline
\multicolumn{13}{l}{\textbf{Non-contextual matching}}  \\ \hline
\multicolumn{1}{l|}{BM25-QR}                    & 0.2          & 0.7          & 1.3           & 2.4           & 214.2           & 219.2          & 60.3          & \multicolumn{1}{c|}{52.9}          & 202.8           & 214.5          & 70.4          & 62.7          \\
\multicolumn{1}{l|}{BE-QR}                    & 0.2          & 0.8          & 1.5           & 2.6           & 207.2           & 213.4          & 72.8          & \multicolumn{1}{c|}{67.2}          & 198.1           & 206.5          & 78.2          & 71.4          \\  \hline
\end{tabular}}
\caption{Automated evaluation metrics on Reddit test set. For MC and SC test set, we both report Perplexity@1/20 and Relevance@1/20; for SC test set, we additionally report Coverage@1/20/100/500. For Coverage@K and Relevance@K, we report the numerator of its percentage, and the larger the better; for Perplexity@K, the smaller the better.}
\label{tab:reddit-result}
\end{table*}

\begin{table}[]
\centering
\scalebox{0.85}{
\begin{tabular}{ccccc} \toprule
\multirow{2}{*}{Retriever} & \multicolumn{4}{c}{Coverage@K}     \\ 
                           & Top-1 & Top-20 & Top-100 & Top-500 \\  \midrule
BM25-QC                    & 16.2  & 28.5   & 35.7    & 42.9    \\
BM25-QS                    & 16.3  & 28.3   & 35.1    & 42.8    \\ \hline
BE-QC                      & 19.6  & 36.2   & 46.4    & 56.5    \\
BE-QS                      & 22.1  & 38.9   & 49.7    & 60.2    \\
TE-DQS                     & 21.5  & 38.4   & 49.5    & 60.4    \\ \hline
CFC-QC                  & 24.2  & 39.1   & 48.6    & 58.2    \\
CFC-QS                  & \textbf{28.8}  & \textbf{43.7}   & \textbf{52.8}    & \textbf{62.6}    \\
CFC-DQS                 & 28.2  & 43.3   & 52.5    & 61.9   \\ \bottomrule
\end{tabular}}
\caption{Automated evaluation metrics on Twitter test set, we report Coverage@1/20/100/500 on the MC test set.}
\label{tab:twitter-result}
\end{table}

\subsection{Compared Models}

For baselines, we select BM25 \cite{robertson2009probabilistic} as sparse representations based method, which is widely used in real scenarios in text matching. Based on BM25 system and the two matching methods (QC and QS matching), two retrievers can be obtained, denoted as BM25-QC and BM25-QS respectively. We choose multi-tower models as dense representations based methods. They are \textbf{b}i-\textbf{e}ncoder based two-tower models for QC matching and QS matching (denoted as BE-QC and BE-QS), and \textbf{t}ri-\textbf{e}ncoder based three-tower model for DQS matching (denoted as TE-DQS). In addition, to demonstrate the advantages of contextual matching, we also report the results of query-response (QR) matching, two retrievers are build based on BM25 system and two-tower model (denoted as BM-QR and BE-QR). 


There are three variants of our proposed CFC models, they are the distilled versions of BE-QC, BE-QS and TE-DQS, which are called CFC-QC, CFC-QS and CFC-DQS respectively. The distillation of each student model needs to train the corresponding teacher model. In particular, the distillation from TE-DQS to CFC-DQS requires two teacher models, because the similarity between both query-context and query-response needs to be calculated. 

We summarize the details of compared models and provide training details in Appendix \ref{section:model_detail}.







\subsection{Evaluation Metrics}
\label{section:metric}


Following previous work \cite{xiong2020approximate, karpukhin2020dense}, \textbf{Coverage@K} is used to evaluate whether Top-K retrieved candidates include the ground-truth response. It is equivalent to recall metric $R_{M}@K$ that often used in fine-grained RS, where $N$ is the size of candidate database. However, Coverage@K is only suitable for evaluating the MC test set, and it is incapable for evaluating the overall retrieval quality due to the one-to-many relationship between context and response. As a supplement, we propose two automated evaluation metrics based on pre-trained models, i.e., \textbf{Perplexity@K} and \textbf{Relevance@K}. For retrieved Top-K responses, DialogGPT \cite{zhang2019dialogpt} is used to calculate the conditional perplexity of the retrieved response given the query. DialogGPT is a language model pre-trained on 147M multi-turn dialogue from Reddit discussion thread and thus very suitable for evaluating our created Reddit dataset. Perplexity@K is the average perplexity of Top-K retrieved responses. In addition to Perplexity, we also evaluate the correlation between the query and retrieved response. We use DialogRPT \cite{gao2020dialogue}, which is pre-trained on large-scale human feedback data with the \emph{human-vs-rand} task that predicts how likely the response is corresponding to the given context rather than a random response. Relevance@K is the average predicted correlation degree between query and Top-K retrieved responses. Perplexity@K and Relevance@K are average metrics based on all Top-K retrieved responses, so they can reflect the overall retrieval quality.

\subsection{Overall Performance}


We demonstrate the main results in Table \ref{tab:reddit-result} and Table \ref{tab:twitter-result} and discuss model performance from multiple perspectives.







\paragraph{Dense vs. sparse} ~ {It can be seen that the performance of dense retrievers far exceed that of the BM25 system, which shows rich semantic information of PLMs and additional training can boost the performance of the retriever. For example, compared with BM25 system, the best undistilled dense retrievers (BE-QS) have a obvious improvement in three metrics. For Coverage@K, the Top-500 recall rate of BE-QS on the MC test set of Reddit and Twitter increase by 12.1\% and 17.4\% absolute compared with BM25-QS. For Perplexity@K, the Top-20 average perplexity of BE-QS on the MC and SC test sets of Reddit is reduced by 8.1 and 8.5 absolute compared with BM25-QS. For Relevance@K, the Top-20 average relevance of BE-QS on the MC and SC test sets on Reddit increase by 6.3\% and 6.5\% absolute compared with BM25-QS. Coverage@K measures the retriever's ability to retrieve gold response, while Perplexity@K and Relevance@K measure the overall retrieval quality. Our results show the consistency of the three metrics, namely, the recall rate and the overall retrieval quality have a positive correlation. 

 




}



\paragraph{Matching method} ~ {Compared with contextual matching, query-response (QR) matching has a much lower retrieval recall rate, which is also verified in \cite{lan2020ultra}. We think it is because that response is usually a short text of one-sentence and contains insufficient information, and there may be little keywords that overlap with the query. Therefore, it is important to consider contextual matching in the RBD system.  

Compared to QC matching, QS and DQS matching should be encouraged in practice due to the additional information provided by the response. However, the BM25 system can not make good use of the information of response, as BM25-QS model does not show obvious advantages over BM25-QC on both Reddit and Twitter datasets. In contrast, dense retrieval models can effectively utilize the response. For example, BE-QS outperforms BE-QC greatly by 7.9\% absolute in terms of Top-500 response retrieval recall rate in MC test set of Reddit. For QS and DQS matching, there is little difference in performance. Especially for SC test set on Reddit and MC test set on Twitter, the performance difference is minimal. One potential advantage of DQS is that it can utilize positive query-response pairs, whose number is much larger than positive query-context pairs. 






}


\paragraph{Distillation benefit}\label{sec:benefit} ~ {We further focus on the performance gain from fine-to-coarse distillation. The distilled models achieve obvious improvement in all three metrics. An obvious pattern is that the distilled models get more larger improvement with a smaller K. Take Twitter dataset as example, the Top-500 retrieval recall rate of CFC models increase by 1.5$\sim$2.4 after distillation, while the Top-1 retrieval recall rate increased by 4.6$\sim$6.7. On Perplexity@K and Relevance@K, our CFC models has similar performance. The significant improvement in the retrieval recall rate at small K’s is especially beneficial to fine-grained response selection, because it opens up more possibility to the ranker to choose good response while seeing fewer candidates. The above results indicate that our student models benefit from learning or inheriting fine-grained knowledge from teacher models. To more clearly demonstrate the performance gains of our model after distillation, we provide the specific values of these gains in Table \ref{tab:performance_gain} in Appendix \ref{section:benefit}.}

\paragraph{Difference between Reddit and Twitter}\label{twitter} ~ {Since DialogGPT and DialogRPT is not pre-trained on Twitter, Perplexity@K and Relevance@K are not suitable for evaluating Twitter dataset. Therefore, we do not build SC test set for Twitter. Compared to Twitter, the Reddit dataset we use is much larger with more common multi-turn conversations, and significantly higher retrieval difficulty. The Top-500 retrieval recall rate on Twitter reach 60\%, while Reddit only reached about 20\%, which indicates that the coarse-grained response retrieval task in open domain conversations still has great challenges.}


\section{Further Analysis}

\subsection{Parameter Sharing}

\begin{table}
\centering
\scalebox{0.9}{
\begin{tabular}{ccccc} \toprule
\multicolumn{1}{c}{\multirow{2}{*}{Retriever}} & \multicolumn{4}{c}{Coverage@K}     \\
\multicolumn{1}{c}{}                           & Top-1 & Top-20 & Top-100 & Top-500 \\ \midrule
BE-QC                                          & 1.31  & 5.28   & 8.12    & 12.26   \\
$\hookrightarrow$ share                               & 1.29  & 5.26   & 8.12    & 12.26   \\ \hline \hline
TE-DQS                                         & 1.47  & \textbf{5.52}   & \textbf{9.74} & \textbf{18.12}   \\
$\hookrightarrow$ share                                     & \textbf{1.49}  & 5.51   & 9.73    & 18.11  \\ \bottomrule
\end{tabular}}
\caption{Impact of parameter sharing on model performance.}
\label{tab:share-parameters}
\end{table}

\begin{figure}
\centering
\includegraphics[width=0.46\textwidth]{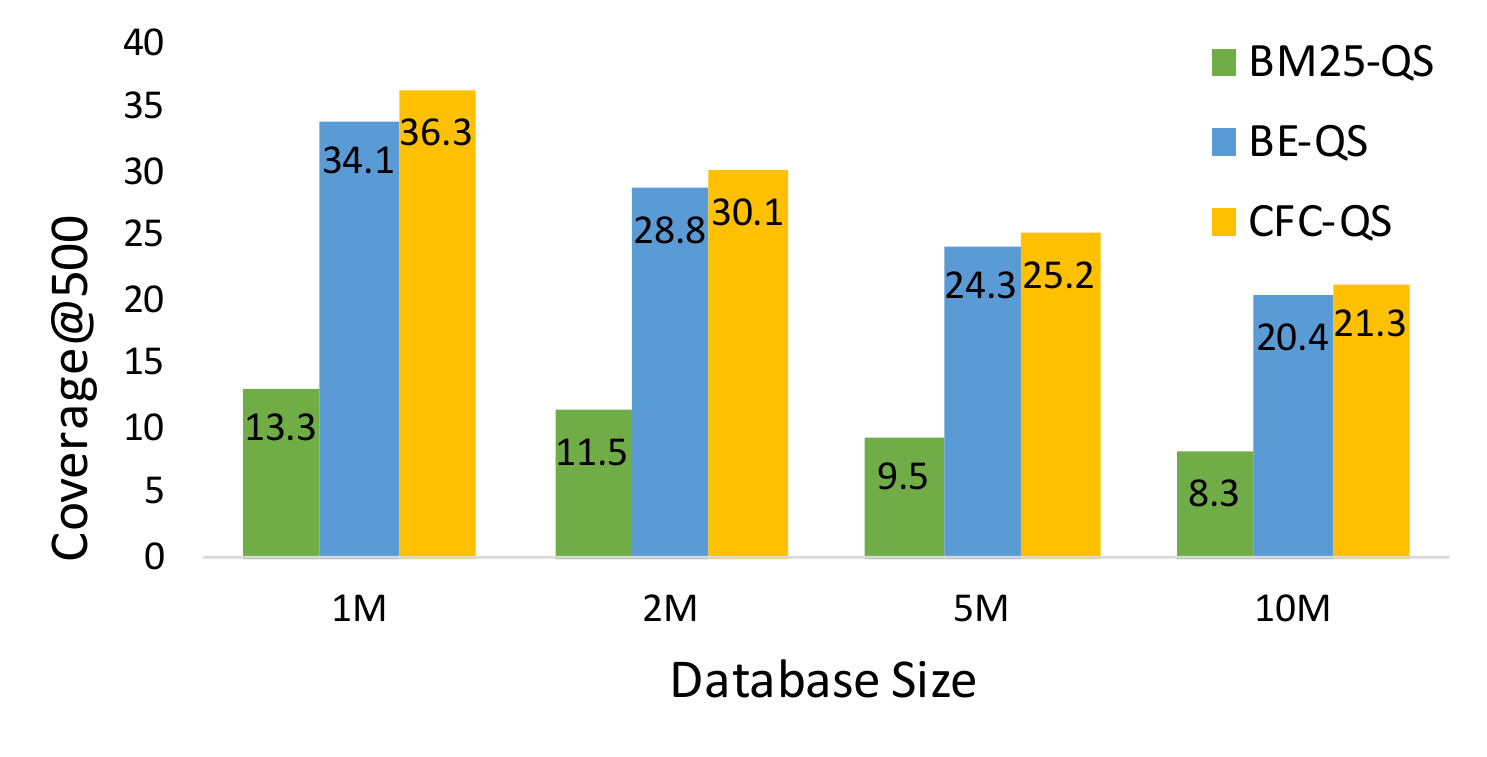} 
\caption{The Impact of database size on Coverage@500 metric of BM25-QS, BE-QS, CFC-QS.} 
\label{fig:database}
\end{figure}

Sharing parameters in dual-encoder structure is a common practice. As shown in Figure \ref{fig:model}, for the encoders in the dotted line, sharing parameters may be beneficial. We try parameter sharing settings on the BE-QC and TE-DQS models, respectively. We add two sets of experiments on the MC test set of Reddit, as shown in Table \ref{tab:share-parameters}. The results show that whether or not to share parameters has little impact on Coverage@K. Therefore, we can share encoder parameters to reduce model complexity with little loss of performance.

Our guess is as follows, the sampling strategy (with replacement) create a certain probability that the query and the context are exactly the same, so the multi-tower model can learn that two identical samples are positive samples for each other, even if the parameters of the encoders are not shared.

\subsection{Effect of Database Size}


We discuss the impact of the size of candidate database on the performance of the model. For different candidate database size (from one million to ten million), we compare the Coverage@500 metric of BM25-QS, BE-QS, and CFC-QS on the MC test set of Reddit (Figure \ref{fig:database}). It can be seen that Coverage@500 shows a slow downward trend as the database size increases. Increasing the size of the database will not make the model performance drop rapidly, which shows the effectiveness and robustness of our models.

\subsection{Human Evaluation}

\begin{table}
\centering
\scalebox{0.9}{
\begin{tabular}{ccc} \hline
        & Avg. Rank & Cohen's Kappa \\ \hline
CFC-QS  & 1.448     &    0.728           \\
BE-QS   & 2.056     &    0.647        \\
BM25-QS & 2.494     &    0.626          \\ \hline  
\end{tabular}}
\caption{Human average rank score of BM25-QS, BE-QS and CFC-QS.}
\label{tab:human_eval_rank}
\end{table}

\begin{table}
\centering
\scalebox{0.8}{
\begin{tabular}{lccc} \hline
                    & \multicolumn{1}{l}{Win} & Loss  & Cohen's Kappa \\ \hline
CFC-QS vs. BE-QS & 0.747                   & 0.253 &     0.634          \\
CFC-QS vs. BM25-QS & 0.816                   & 0.184 &     0.672   \\ \hline    
\end{tabular}}
\caption{Human pairwise comparison of BM25-QS, BE-QS and CFC-QS.}
\label{tab:human_eval_compare}
\end{table}

To further evaluate and compare our models, we conduct a human evaluation experiment. We random select 1000 queries from the MC and SC test set (500 each) of Reddit dataset, and retrieve the Top-1 response by the BM25-QS, BE-QS and CFC-QS models respectively. Three crowd-sourcing workers are asked to score the responses. For each query, the annotator will strictly rank the retrieved responses of the three models. We report the average rank scores (between 1 and 3, the smaller the better) and the winning rate in pairwise comparison. Each two annotators have a certain number (about 200) of overlapping annotated samples. To evaluate the inter-rater reliability, the Cohen's kappa coefficient \cite{kraemer2014kappa} is adopted. 

Table \ref{tab:human_eval_rank} and Table \ref{tab:human_eval_compare} report the average ranking score of each model and pairwise comparison between models respectively. The average ranking score of CFC-QS is the highest, and CFC-QS can beat BE-QS and BM25 in most cases (74.7\%$\sim$81.6\%), which indicates CFC-QS occupies a clear advantage in Top-1 retrieval. All Cohen’s Kappa coefficients is between 0.6 and 0.7, indicating annotators reach moderate agreement. The results of human evaluation further verify the performance improvement brought by distillation to the model. We select several examples with human evaluation as case study and these results are presented in Appendix \ref{section:case}.

\subsection{Retrieval efficiency}

We compare the retrieval latency of BM25-QS and BE-QS on the reddit MC test set, which represent the efficiency of the sparse and dense retriever respectively. We fix the batch size to 32 and retrieve top 100 most similar candidates. With the help of FAISS index, the average retrieval time of each batch by BE-QS is 581.8ms. In contrast, the average retrieval time by BM25 system using file index is 1882.6ms, about three times that of BE-QS. This indicates that the dense retriever also has an advantage in retrieval efficiency.




The relatively inferior of dense retriever is that it needs to compute the embeddings of the candidate database and establish the FAISS index, which is quite time-consuming and it takes about 9 hours for BE-QS to handle 10 million candidates with 8 GPUs, while it only takes about 10 minutes to build a BM25 index.  

Since distillation does not change the structure of the retriever, it will not affect the retrieval efficiency. The cost of distillation is mainly reflected in the training of the teacher model and the extensive forward calculation in the distillation process.
 
\section{Conclusion}
\label{section:conclusion}

In this paper, we propose a Contextual Fine-to-Coarse (CFC) distilled model. In CFC model, we adopt matching on both query-response and query-context. Considering the retrieval latency, we use multi-tower architecture to learn the \textbf{dense representations} of queries, responses and corresponding contexts. To further enhance the performance of the retriever, we distill the knowledge learned by the one-tower architecture (fine-grained) into the multi-tower architecture (coarse-grained). We construct two new datasets based on Reddit comment dump and Twitter corpus, and extensive experimental results demonstrate the effectiveness and potential of our proposed model. In the future work, we will further explore how the enhancement of coarse-grained RS can help fine-grained RS.

\section*{Acknowledgments}

This work is partially supported by Natural Science Foundation of China (No.6217020551, No.61906176), Science and Technology Commission of Shanghai Municipality Grant (No.20dz1200600, 21QA1400600, GWV-1.1, 21511101000) and Zhejiang Lab (No. 2019KD0AD01).

\section*{Ethical Statement}

In this paper, different ethical restrictions deserve discussion.

The datasets we created are derived from large dialogue corpus that publicly available on the Internet, and we strictly followed the platform's policies and rules when obtaining data from web platforms. We did not use any author-specific information in our research.

Online large dialogue corpus may includes some bias, such as political bias and social bias, and our model might have inherited some forms of these bias. In order to limit these bias as much as possible, we filter controversial articles and removed data with offensive information when possible.

\bibliography{anthology,custom}
\bibliographystyle{acl_natbib}

\appendix

\section{Dataset Construction Details}
\label{appendix:data_detail}


To filter boring and dull content and speed up the retrieval speed, we set a limit for the length of contexts and responses. We limit the context to contain at least 5 words and less than 128 words, and the response contains at least 5 words and less than 64 words. It is specially beneficial to limit the length of the response, since according to our statistics, many short responses such as "\emph{Fair Enough}" and "\emph{Thanks :D}" may have large number (\emph{tens of thousands}) of different contexts. 

Besides, we also limit the upper limit of the number of contexts corresponding to the response. The number of contexts of each response in the MC test set is limited to no more than 50, which is to prevent the selected responses from being a meaningless universal response. The detailed construction of the two test sets is described in Algorithm \ref{algorithm:test_set}.

\begin{algorithm}
\caption{Construction of SC \& MC test set.}
\label{algorithm:test_set}
\begin{algorithmic}[1]
\State{$R$: A set of unique responses.}
\State{$SC' = \varnothing$}
\State{$MC' = \varnothing$}
\ForEach{$r \in R$}%
\State{$C_{r}=\textrm{FindAllContexts}(r)$}  \algorithmiccomment{Find all contexts whose response is $r$.}
\If {$|C_{r}| > 1$} 
\State{$C_{r}^{-}, c = \textrm{Split}(C_{r})$} \algorithmiccomment{Random pick one context $c$ from $C_{r}$, the remaining contexts is denoted as $C_{r}^{-}$.}
\State{$MC' = MC' \cup \{c, r\}$} 
\Else
\State{$SC' = SC' \cup \{c \in C_{r}, r\}$} 
\EndIf
\EndForEach
\State{$MC = \textrm{RandomSample}(MC')$} 
\State{$SC = \textrm{RandomSample}(SC')$} 
\State \Return{$SC$, $MC$}
\end{algorithmic}
\end{algorithm}


To construct the training set, we need to find out responses that corresponding multiple contexts. We use dict to implement it, where the key is the response and the value is the list of corresponding contexts. During the training of the multi-tower model, in each iteration, a batch of keys is randomly sampled from the dict. For each key (i.e., each response) in the batch, two contexts are randomly selected from the corresponding value (i.e., the list of contexts), one of which is used as the query and the other is used as a positive context, and the key is used as a positive response. The other contexts and responses in the batch are all negative instances of the query.




\begin{table*}
\centering
\scalebox{0.75}{
\begin{tabular}{c|l|clclc}
\hline
\multirow{2}{*}{\textbf{Match}} & \multirow{2}{*}{\textbf{Model-ID}} & \multicolumn{1}{c|}{\multirow{2}{*}{\textbf{Architecture}}} & \multicolumn{2}{c|}{\textbf{Training}}                                                                 & \multicolumn{2}{c}{\textbf{Inference}}               \\ \cline{4-7} 
                                  &                                 & \multicolumn{1}{c|}{}                                       & \multicolumn{1}{c|}{\textbf{Input}}                               & \multicolumn{1}{c|}{\textbf{Loss}} & \multicolumn{1}{c|}{\textbf{Input}} & \multicolumn{1}{c}{\textbf{Output}}  \\ \hline
\multirow{2}{*}{QC}                & BE-QC(S)                      & \multicolumn{1}{c|}{Two-Tower}                              & \multicolumn{1}{l|}{QY, POS CXT, NEG CXTs}                        & \multicolumn{1}{c|}{CT}          & \multicolumn{1}{l|}{QY, CXT}        & \multicolumn{1}{c}{DSS}              \\ 
                                  & BE-QC(T)                          & \multicolumn{1}{c|}{One-Tower}                              & \multicolumn{1}{l|}{QY, CXT, LABEL}                               & \multicolumn{1}{c|}{CE}            & \multicolumn{1}{l|}{QY, CXT}        & \multicolumn{1}{c}{FSS}              \\ \hline
\multirow{2}{*}{QS}                & BE-QS(S)                       & \multicolumn{1}{c|}{Two-Tower}                              & \multicolumn{1}{l|}{QY, POS SESS, NEG SESSs}                      & \multicolumn{1}{c|}{CT}          & \multicolumn{1}{l|}{QY, SESS}       & \multicolumn{1}{c}{DSS}              \\  
                                  & BE-QS(T)                          & \multicolumn{1}{c|}{One-Tower}                              & \multicolumn{1}{l|}{QY, SESS, LABEL}                              & \multicolumn{1}{c|}{CE}            & \multicolumn{1}{l|}{QY, SESS}       & \multicolumn{1}{c}{FSS}              \\ \hline
\multirow{3}{*}{DQS}               & TE-DQS(S)                       & \multicolumn{1}{c|}{Three-Tower}                              & \multicolumn{1}{l|}{QY, POS CXT, NEG CXTs, POS RESP,   NEG RESPs} & \multicolumn{1}{c|}{CT}          & \multicolumn{1}{l|}{QY, CXT, RESP}  & \multicolumn{1}{c}{DSS}              \\  
                                  & TE-DQS(T1)                       & \multicolumn{1}{c|}{One-Tower}                              & \multicolumn{1}{l|}{QY, CXT, LABEL}                               & \multicolumn{1}{c|}{CE}            & \multicolumn{1}{l|}{QY, CXT}        & \multicolumn{1}{c}{FSS}              \\  
                                  & TE-DQS(T2)                      & \multicolumn{1}{c|}{One-Tower}                              & \multicolumn{1}{l|}{QY, RESP, LABEL}                              & \multicolumn{1}{c|}{CE}            & \multicolumn{1}{l|}{QY, RESP}       & \multicolumn{1}{c}{FSS}              \\ \hline
QR       & BE-QR        & \multicolumn{1}{c|}{Two-Tower} & \multicolumn{1}{l|}{QY, POS RESP, NEG RESPs} & \multicolumn{1}{c|}{CE} & \multicolumn{1}{l|}{QY, RESP} & DSS \\ \hline
\multicolumn{7}{l}{\textbf{Abbreviation}}   \\ 
\multicolumn{7}{p{20cm}}{S(Student), T(Teacher), QY(Query), CXT(Context), RESP(Response), SESS(Session), POS(Positive), NEG (Negative), CT(Contrastive), CE(Cross Entropy), DSS(Dot-product based Similarity Score), FSS(Feature based Similarity Score)}  \\ \hline
\end{tabular}}
\caption{The input, output and training objectives of tower models in this paper. For each matching method, one or two teacher models need to be trained for knowledge distillation.}
\label{tab:arch}
\end{table*}

\begin{table*}
\centering
\scalebox{0.85}{
\begin{tabular}{clclcccccccc} \toprule
\multirow{2}{*}{Dataset} & \multicolumn{3}{c}{Distillation} & \multicolumn{4}{c}{Coverage@K}     & \multicolumn{2}{c}{Perplexity@K} & \multicolumn{2}{c}{Relevance@K} \\
                         & Before   &   & After           & Top-1 & Top-20 & Top-100 & Top-500 & Top-1          & Top-20          & Top-1          & Top-20         \\ \midrule
\multirow{3}{*}{Reddit}   & BE-QC          & $\dashrightarrow$ & CFC-QC          & +1.6  & +1.2   & +1.0    & +0.7    & -5.9           & -2.6            & +3.6           & +2.7           \\
                         & BE-QS          & $\dashrightarrow$ & CFC-QS          & +2.6  & +1.9   & +1.3    & +0.9    & -5.3           & -3.0            & +2.8           & +2.7           \\
                         & TE-DQS         & $\dashrightarrow$ & CFC-DQS         & +2.3  & +1.8   & +2.9    & +1.3    & -4.9           & -2.1            & +2.1           & +2.1           \\ \hline
\multirow{3}{*}{Twitter} & BE-QC          & $\dashrightarrow$ & CFC-QC          & +4.6  & +2.9   & +2.2    & +1.7    & -              & -               & -              & -              \\
                         & BE-QS          & $\dashrightarrow$ & CFC-QS          & +6.7  & +4.8   & +3.1    & +2.4    & -              & -               & -              & -              \\
                         & TE-DQS         & $\dashrightarrow$ & CFC-DQS         & +6.7  & +4.9   & +3.0    & +1.5    & -              & -               & -              & -              \\ \bottomrule
\end{tabular}}
\caption{Model performance gain after distillation on the MC test set of Reddit and Twitter dataset.}
\label{tab:performance_gain}
\end{table*}

\begin{table*}
\centering
\scalebox{0.73}{
\begin{tabular}{c|c|c|l|c}
\hline
Number                  & Query                                                         & Method & \multicolumn{1}{c|}{Response}                                                                                                                                                                          & \multicolumn{1}{l}{Rank} \\ \hline
\multirow{4}{*}{Case 1} & \multirow{4}{*}{My pc   Isn't good enough unfortunately}      & Gold   & How old is your   computer ?                                                                                                                                                                           & -                              \\ \cline{3-5} 
                        &                                                               & CFC-QS & what are your PC   specs ?                                                                                                                                                                             & 1                              \\ \cline{3-5} 
                        &                                                               & BE-QS  & \begin{tabular}[c]{@{}l@{}}Idk but apps   aren't great on ps4 .\\ My roku ultra is much faster for whatever reason .\end{tabular}                                                                      & 2                              \\ \cline{3-5} 
                        &                                                               & BM25   & \begin{tabular}[c]{@{}l@{}}I' m on the edge   . deals are good , \\ but good enough to reactivate my pc ?\end{tabular}                                                                                 & 3                              \\ \hline
\multirow{4}{*}{Case 2} & \multirow{4}{*}{Can I   get Spider Man 2099}                  & Gold   & Good trade ,   thanks !                                                                                                                                                                                & -                              \\ \cline{3-5} 
                        &                                                               & CFC-QS & You got it PM   sent !                                                                                                                                                                                 & 1                              \\ \cline{3-5} 
                        &                                                               & BE-QS  & Sure , 1 by   Paypal pls :  xxx@hotmail.com                                                                                                                                                            & 2                              \\ \cline{3-5} 
                        &                                                               & BM25   & \begin{tabular}[c]{@{}l@{}}right now ,   Spider man 2099 is \\ the best written spider man .\end{tabular}                                                                                              & 3                              \\ \hline
\multirow{4}{*}{Case 3} & \multirow{4}{*}{Gut   Knife Scorched FT , worth 19keys}       & Gold   & No thanks .   Sorry                                                                                                                                                                                    & -                              \\ \cline{3-5} 
                        &                                                               & CFC-QS & I only have   15keys .                                                                                                                                                                                 & 1                              \\ \cline{3-5} 
                        &                                                               & BE-QS  & Add me on steam   ! Nvm I added you .                                                                                                                                                                  & 2                              \\ \cline{3-5} 
                        &                                                               & BM25   & Nah only keys ,   knives are meh to me , all of'em .                                                                                                                                                   & 3                              \\ \hline
\multirow{4}{*}{Case 4} & \multirow{4}{*}{The   email is returning failures to deliver} & Gold   & Should be   working now .                                                                                                                                                                              & -                              \\ \cline{3-5} 
                        &                                                               & CFC-QS & THE email ? It's   just email ! !                                                                                                                                                                      & 3                              \\ \cline{3-5} 
                        &                                                               & BE-QS  & \begin{tabular}[c]{@{}l@{}}It asks for your   username I think , doesn't it ? \\ Try just enter your username you used to register\\ instead of the email and let me know if that works .\end{tabular} & 1                              \\ \cline{3-5} 
                        &                                                               & BM25   & did you get my   email with the pic ?                                                                                                                                                                  & 2                              \\ \hline
\end{tabular}
}
\caption{Four retrieved cases on our human evaluation set. We report Top-1 retrieved response of the three models as well as gold response. The Rank column is the ranking of the three responses given by the annotator (the lower the better).}
\label{tab:case_study}
\end{table*}

\section{Model Details} 
\label{section:model_detail}


Due to the different matching methods, the training of different retrievers requires slightly different input. Taking BE-QC as an example, given a query, positive and negative contexts are needed to learn the representation of query and contexts, while in BE-QS, positive and negative sessions are required. Besides, the distillation of each student model requires training corresponding teacher model, and the data of training teacher model is consistent with the student model. We summarize the input, output, and training objectives of student and teacher models in Table \ref{tab:arch}.


To implement the BM25 method, we use Elasticsearch\footnote{\url{https://www.elastic.co/}}, which is a powerful search engine based on Lucene library \cite{bialecki2012apache}. For dense retrieval methods, FAISS \cite{johnson2019billion} toolkit is used to retrieve candidate vectors. All encoders in our tower models (including one-tower, two-tower and three-tower) are initialized with \emph{bert-base}\footnote{\url{https://huggingface.co/bert-base-uncased}}, which includes 12 encoder layers, embedding size of 768 and 12 attention heads. For dense models (BE-QC, BE-QS, TE-DQS), we use the same batch size of 32 for Reddit and Twitter, and we train 30 epochs on Reddit and 10 epochs on Twitter. For all teacher models, we use the same batch size of 16, and we train 40 epochs on Reddit and 20 epochs on Twitter. For the distillation (CFC-QC, CFC-QS, CFC-DQS), we train additional 10 epochs on reddit and 5 epochs on twitter respectively, starting from the early checkpoints (20 epochs in Reddit and 5 epochs in Twitter for fair comparison) of BE-QC, BE-QS, TE-DQS. We use Adam \cite{kingma2014adam} optimizer with learning rate of 2e-4 and the warmup steps of 200 to optimize the parameters. We set the knowledge distillation temperature to 3 and the rate of distillation loss to 1.0. All experiments are performed on a server with 4 NVIDIA Tesla V100 32G GPUs.

\section{Distillation Benefit}
\label{section:benefit}

To more clearly show the performance gains of our model after distillation, we present the specific values of these gains in Table \ref{tab:performance_gain}. Readers can compare the results in this table when reading the Distillation Benefit part in \S~\ref{sec:benefit}. Positive Coverage@K and Relevance@K, and negative Perplexity@K all represent the improvement of model performance. After the distillation, the accuracy and correlation between the retrieved responses and the query increase, and the conditional perplexity decreases, indicating the huge benefits of distillation.

\section{Case Study}
\label{section:case}

As sparse representations base method, BM25 system tends to retrieve responses that overlaps with the context. For some complicated cases, BM25 cannot correctly retrieve those seemingly unrelated, but are the best answer in the current context.

In second case of Table \ref{tab:case_study}, BM25 selects the response that contains "Spider Man 2099" in the query. But in the context of the forum, "Can I get Spider Man 2099" is actually looking for the e-book files of this comic.  Compared to the comments of Spider Man 2099 given by BM25, our model retrieves "You got it PM (private message) sent!" is a harder to find, but more accurate response.

The third case is an in-game item trading query. In related forums, "keys" are used as currency. "Knife Scorched FT" and "19keys" in query respectively represent an item to be sold and its expected price. The result of BM25 covers "knife" and "key", but the meaning of the whole sentence does not match the query. On the other hand, our model selected "I only have 15keys", a standard bargaining, perfectly match the query. 

There are also some examples such as case 4. Our model gives worse results than BM25. In case 4, CFC-QS retrieves a worse result, and the response retrieved by BE-QS is relatively better. 

\end{document}